\documentclass[preprint]{elsarticle}
\usepackage[utf8]{inputenc}
\usepackage{amssymb}
\usepackage{amsmath,amsthm}
\usepackage{amsfonts}
\usepackage[ruled,vlined]{algorithm2e}
\usepackage{times}
\usepackage{array}
\usepackage{subfig}
\usepackage{hyperref}
\usepackage{url}
\usepackage{pdfpages}
\usepackage{footnote}
\usepackage{array,multirow}
\usepackage{booktabs}
\usepackage{textcomp}

\newtheorem{df}{Definition}[section]

\newcommand{\at}{\makeatletter @\makeatother}

\journal{Expert Systems with Applications}
\begin{document}
\begin{frontmatter}
  \title{Topic Discovery in Massive Text Corpora Based on Min-Hashing\footnote{\textcopyright 2019. Licensed under the Creative Commons \href{https://creativecommons.org/licenses/by-nc-nd/4.0/}{CC-BY-NC-ND 4.0}}}

  \author[mymainaddress]{Gibran Fuentes-Pineda\corref{mycorrespondingauthor}}
  \cortext[mycorrespondingauthor]{Corresponding author}
  \ead{gibranfp@unam.mx}

  \author[mymainaddress]{Ivan V. Meza-Ruiz}
  \ead{ivanvladimir@turing.iimas.unam.mx}

  \address[mymainaddress]{Instituto de Investigaciones en Matemáticas Aplicadas y en Sistemas, Universidad Nacional Autónoma de México, Circuito Escolar s/n 4to piso, Ciudad Universitaria, Coyoacán, 04510, CDMX, Mexico.}

  \begin{abstract}
    Topics have proved to be a valuable source of information for exploring, discovering, searching and representing the contents of text corpora.    
    They have also been useful for different natural language processing tasks such as text classification, text summarization and machine translation.
    Most existing topic discovery approaches require the number of topics to be provided beforehand. However, an appropriate number of topics for a given corpus depends on its characteristics and is often difficult to estimate.
    In addition, in order to handle massive amounts of text documents, the vocabulary must be reduced considerably and large computer clusters and/or GPUs are typically required.
    This paper describes Sampled Min-Hashing (SMH), a scalable approach to topic discovery which does not require the number of topics to be specified in advance and can handle massive text corpora and large vocabularies using modest computer resources. 
    The basic idea behind SMH is to generate multiple random partitions of the corpus vocabulary to find sets of highly co-occurring words, which are then clustered to produce the final topics.
    An extensive qualitative and quantitative evaluation on the 20 Newsgroups, Reuters, Spanish Wikipedia and English Wikipedia corpora shows that SMH is able to consistently discover meaningful and coherent topics at scale. 
    Remarkably, the time required by SMH grows linearly with the size of the corpus and the number of words in the vocabulary; a non-parallel implementation of SMH was able to discover topics from the whole English version of Wikipedia (5M documents approximately) with a vocabulary of 1M words in less than 7 hours.
    Our findings provide further evidence of the relevance and generality of beyond-pairwise co-occurrences for pattern discovery on large-scale discrete data, which opens the door for other applications and several interesting research directions.
\end{abstract}

\end{frontmatter}

\section{Introduction}
Topics are structures that capture the themes present in a collection of text documents.
They have proved to be a valuable source of information for exploring, discovering and searching the contents of text corpora. 
The automatic extraction of these structures from the vector space model has been a challenging and widely studied problem for several decades.
This capability has become essential for many intelligent systems dealing with collections of text documents. 
For example, the field of Digital Humanities makes extensive use of topic discovery tools to gain insights into a corpus~\citep{tmdh} and enable exploratory search~\citep{marchionini2006exploratory, meeks2012digital}, where the discovered topics serve as a guide to further learning about an unfamiliar domain.
A similar practice has been adopted in journalism~\citep{journalism, gunther2016word, guo} and policy-making~\citep{talley2011database, shirota, policy, nichols2014topic, moschella}, among other fields. 

Moreover, topics have been used to obtain a compact representation of documents for classification and retrieval.
The thematic information provided by topics has been useful for a variety of natural language processing tasks such as text classification~\citep{rubin2012statistical}, text summarization~\citep{summ} and machine translation~\citep{mimno2009polylingual}.
These structures have also helped improve or enable several applications of intelligent systems such as hashtag recommendation~\citep{hashtagLDA2016}, online community detection~\citep{communityLDA}, recommender systems~\citep{recommendationLDA2017,recommenderLDA2017}, depression detection~\citep{depressionLDA2015}, link prediction~\citep{tweetLDA2012}, and crime prediction~\citep{crimeLDA2012}.

Latent Dirichlet Allocation (LDA)~\citep{lda} and other topic models have been the predominant topic discovery approaches for over a decade.
Most of these approaches require the number of topics to be provided beforehand. 
However, the appropriate number of topics for a given corpus depends on its characteristics and it is often difficult to estimate.
Unfortunately, both topic quality and discovery time are sensitive to this number.
In addition, in order to handle massive text corpora, the vocabulary must be reduced considerably and large computer clusters and/or GPUs are typically required.

In this paper we present Sampled Min-Hashing (SMH), an alternative approach to topic discovery which builds upon previous work on object discovery from large-scale image collections~\citep{fuentes}. 
The basic idea behind SMH is to generate multiple random partitions of the corpus vocabulary by applying Min-Hashing on the word occurrence space spanned by inverted file lists to find sets of highly co-occurring words, which are then clustered to produce the final topics.
Thanks to the use of Min-Hashing to efficiently find and cluster highly co-occurring word sets, the proposed approach scales well both with the size of the corpora and the number of words in the vocabulary. 
Moreover, since SMH relies on agglomerative clustering, the number of topics is determined by its parameter setting and word co-occurrence characteristics of the corpus, rather than being specified beforehand.
As opposed to LDA and other topic models, where topics are distributions over the words in the vocabulary, SMH topics are sets of highly co-occurring words.
We show that SMH can consistently discover meaningful and coherent topics from corpora of different domains and sizes.

The main contributions of this paper are threefold. First, we describe an algorithm to efficiently mine beyond-pairwise relationships in a collection of bags with integer multiplicities. Second, we introduce a novel topic discovery approach which can handle massive text corpora with large vocabularies using modest computer resources and does not require the number of topics to be provided beforehand. 
Third, we present an extensive evaluation and analysis of the impact of different parameters on the coherence of the discovered topics and the efficiency of SMH.
Preliminary results with an earlier implementation of the proposed approach were presented in~\citep{swmh}. 
The current manuscript provides a more detailed description of the proposed approach, emphasizing the importance of the consistency property of Min-Hashing to mine beyond-pairwise co-occurrences. 
It also presents a more suitable and extensive evaluation based on the methodology proposed by~\citet{npmi}, as well as a more direct comparison between SMH and Online LDA.
In addition, topics were discovered from the Spanish edition of Wikipedia in order to validate the effectiveness of the proposed approach in other languages. 

The remainder of the paper is organized as follows. 
In Sect.~\ref{sec:prevwork}, we review some related work on Min-Hashing and beyond-pairwise relationship mining. 
Section~\ref{sec:minhash} describes the original Min-Hashing scheme for pairwise similarity. 
SMH is presented in detail in Sect.~\ref{sec:smh}. 
The experimental evaluation of the coherence of the discovered topics and the scalability of the approach are reported in Sect.~\ref{sec:eval}. 
Finally, Sect.~\ref{sec:conclu} concludes with some remarks and future work.

\section{Related Work}
\label{sec:prevwork}
\subsection{Min-Hashing}
Locality Sensitive Hashing (LSH) is a randomized algorithm for performing approximate similarity search in high dimensional spaces.
The general idea of LSH is to define a suitable family of similarity-preserving hash functions for randomly projecting the high dimensional space onto a lower dimensional subspace such that the distances between items are approximately preserved. 
Originally, LSH was proposed for efficient pairwise similarity search on large-scale datasets.
However, it has also been used for other purposes. 
For example, to compute a fast proposal distribution when sampling mixtures of exponential families~\citep{fastex}, to efficiently find high-confidence association rules without support pruning~\citep{rulemine}, to retrieve inner products in collaborative filtering~\citep{alsh}, or to accelerate deep neural networks~\citep{dl_lsh}. 

Multiple LSH schemes have been proposed for different metric spaces such as the Hamming distance~\citep{lsh}, the Euclidean distance~\citep{euclidean}, and the Jaccard similarity~\citep{minhash,cohen01}. 
In particular, Min-Hashing~\citep{minhash,cohen01}, an LSH scheme to perform similarity search for sets based on the Jaccard similarity, has been of special interest for document and image retrieval applications because documents and images are often represented as sets of words or visual words. 
However, the original Min-Hashing scheme assumes a set representation (i.e. the presence or absence of words/visual words) of documents or images which is not suitable for many applications where the frequency of occurrence is important~\citep{salton,buckley}. 
For this reason, extensions to the original Min-Hashing scheme have been proposed for bags with both integer and real-valued multiplicities~\citep{chum_wminhash, manesse_cws2010, ioffe_ics2010}. 

Although pairwise similarity search is a building block for several applications, some problems require searching higher-order relationships (e.g. estimating multi-way associations among words from a corpus~\citep{li_multiway2007}, clustering collinear points in high-dimensional spaces~\citep{agarwal2005beyond}, or modeling 3D objects for retrieval and recognition~\citep{zhou2006}). 
One of the most important challenges when searching for higher-order relationships is that the number of combinations to consider increases exponentially with the order of the relationship and the total number of elements in the dataset (e.g. the worst-case query time to find the maximum relationship of order $3$ in a dataset of $N$ elements with $D$ dimensions is $O(N^3)$~\citep{shrivastava2013beyond}).
Interestingly, the space partitioning defined by Min-Hashing not only approximately preserves pairwise similarities but also higher order relationships based on the \emph{Jaccard Co-occurrence Coefficient}, an extension of the Jaccard similarity for measuring beyond-pairwise relationships among sets~\citep{fuentes, shrivastava2013beyond}.
Shrivastava and Li~\citep{shrivastava2013beyond} proposed a new bucketing scheme for Min-Hashing in order to perform $k$-way similarity searches, which was applied to finding sets of semantically similar words and enhancing document retrieval with multiple queries.

The ability of Min-Hashing to capture beyond-pairwise relationships has been exploited to mine visual word co-occurrences from a collection of images by applying it to the inverted file lists instead of the bag-of-words representations of images~\citep{chum_wminhash,fuentes}.
In particular, \citep{fuentes} used Min-Hashing to discover objects by treating each cell from each partition as a sample composed of visual word sets that frequently co-occur in the collection, and then clustering overlapping inter-partition cells to form complete objects.
Here, we hypothesize that words frequently occurring together in the same document in a given corpus likely belong to the same topic and we can therefore discover topics by applying Min-Hashing to the inverted file lists of the corpus, which represent word occurrences.
To achieve this, we extend the approach proposed by~\citet{fuentes} to take into account word frequencies, which have shown to be relevant in Natural Language Processing and Information Retrieval tasks.

\subsection{Topic Discovery}
Many different approaches for discovering topics have been proposed in the past few decades, including Latent Semantic Analysis (LSA)~\citep{lsa} and its probabilistic extension~\citep{plsa}, as well as various topic models~\citep{lda, ctm2005, pachinko2006, Ruslan_undirected_nips2009,Srivastava_uai2013,larochelle_nips2012}.
One of the most widely used approaches to date has been Latent Dirichlet Allocation (LDA)~\citep{lda}, a directed graphical model with latent topic variables, where topics are distributions over the words of a given vocabulary and documents are likewise distributions over $K$ topics.
LDA assumes that the number of topics $K$ is specified by the user. 
However, an appropriate number is not easy to deduce by simple inspection and an improper choice may lead to low-quality topics~\citep{stability}. 
To address this problem, the Hierarchical Dirichlet Process (HDP)~\citep{hdp} extends LDA by using Dirichlet processes to model the uncertainty with respect to the number of topics, thus allowing $K$ to be unbounded and automatically inferred from the corpus.
Unfortunately, this comes at the price of greater time complexity.
On the other hand, for both LDA and HDP, the vocabulary is usually a small subset of all words in the corpus, selected by the user so as to improve discovery time and topic quality by considering only the $D$ most frequent words.

A remaining challenge for using LDA, HDP and other topic models is the scalability of the inference process used to extract topics. 
In recent years much research has been devoted to devising more scalable inference methods. 
In particular, several parallel and distributed versions of Markov Chain Monte Carlo samplers have been proposed~\citep{LightLDA2015, SaberLDA2017,NomadLDA2015, BigTopic2017, LDAs2017, HarpLDA2016, WarpLDA2016}.
These samplers have enabled the discovery of thousands and even millions of topics in massive text corpora with large vocabulary sizes but require computer clusters or GPUs to be practical. 
Other alternatives to MCMC samplers include methods based on Variational Bayesian formulations such as Online LDA~\citep{onlinelda, stochasticlda}, HSVG~\citep{Mimno_icml2012} or SCVB0~\citep{SCVI2013}, which can perform online inference for LDA and HDP on massive text corpora with large vocabulary sizes and numbers of topics using modest computer resources. 

We propose a different approach to topic discovery, which is able to efficiently discover topics in massive text corpora with large vocabularies while automatically adapting the number of topics according to the corpus characteristics. 
As opposed to LDA and other topic models, which define a generative process for documents from which topics are found by computing the posterior distribution of the hidden variables given the observed words in the corpus, SMH samples directly from the co-occurrence space of words to identify those belonging to the same topic.
While LDA produces topics which are distributions over a given vocabulary, the topics produced by our approach are subsets of highly co-occurring words.  
In order to validate our approach, we compare it with Online LDA both in terms of efficiency and topic coherence using different corpora.

\section{Min-Hashing for Similarity Search}
\label{sec:minhash}
Min-Hashing is an LSH scheme in which hash functions are defined with the property that the probability of any pair of sets $\{S_i, S_j\}, i,j \in \{1, \ldots N\}$ having the same value is equal to their Jaccard Similarity, i.e.,

\begin{equation}
  \label{eq:mh_prob}
  P[h(S_i) = h(S_j)] = \frac{\mid S_i \cap S_j \mid}{\mid S_i \cup S_j \mid} = J(S_i, S_j) \in [0,1].
\end{equation}

A MinHash function can be implemented as follows. First, a random permutation $\pi$ of all the elements of the universal set $U = \{1, \ldots, D\}$ is generated.
Then, the first element of the sequence $(\pi(S_i)_k)_{k=1}^{\vert S_i \vert }$ induced by $\pi$ on each set $S_i, i = 1, \ldots, N$ is assigned as its MinHash value, that is to say $h(S_i) = \pi(S_i)_1$. 
Since similar sets share many elements they will have a high probability of taking the same MinHash value, whereas dissimilar sets will have a low probability.
Usually, $M$ different MinHash values are computed for each set from $M$ different MinHash functions $h_m, m = 1, \ldots M$ using $M$ independent random permutations.
It has been shown that the proportion of identical MinHash values between two sets from the $M$ independent MinHash functions is an unbiased estimator of their Jaccard similarity~\citep{minhash}.

The original Min-Hashing scheme has been extended to perform similarity search on integer and real-valued bags~\citep{charikar, chum_wminhash, manesse_cws2010, ioffe_ics2010}, generalizing the Jaccard similarity to

\begin{equation}
  J_{B}(B_i, B_j) = \frac{\sum_{w = 1}^{D} \min (B_i^w, B_j^w)}{\sum_{w = 1}^{D} \max(B_i^w, B_j^w)} \in [0,1],
  \label{eq:jb}
\end{equation}

\noindent where $B_i^w$ and $B_j^w$ are the integer or real-valued multiplicities of the element $w$ in the bags $B_i$ and $B_j$, respectively\footnote{Note that Eq.~\ref{eq:jb} reduces to Eq.~\ref{eq:mh_prob} if all multiplicities in bags $B_i$ and $B_j$ are either 0 or 1, i.e. $B_i$ and $B_j$ represent sets, since $\sum_{w = 1}^{D}\min (B_i^w, B_j^w)$ corresponds to counting the number of common elements and $\sum_{w = 1}^{D} \max(B_i^w, B_j^w)$ to counting the number of elements in both bags.}. In particular, \citet{chum_wminhash} proposed a simple strategy for bags with integer-valued multiplicities where each bag $B_i, i = 1, \ldots, N$ is converted to a set $\hat{S}_i$ by replacing the multiplicity $B_i^w$ of the element $w$ in $B_i$ with $B_i^w$ new elements $e_1, \ldots, e_{B_i^w}$. In this way, an extended universal set is created as
\begin{equation*}
  U_{ext} = \{1, \ldots, F_1, \ldots, F_1 + \cdots + F_{D - 1} + 1, \ldots , F_1 + \cdots + F_D \}
\end{equation*}

\noindent where $F_1, \ldots , F_D$ are the maximum multiplicities of elements $1, \ldots, D$, respectively. Thus, the application of the original Min-Hashing scheme described above to the converted bags $\hat{S}_i \subseteq U_{ext}$ adheres to the property that $P[h(\hat{S}_i) = h(\hat{S}_j)] = J_B(B_i, B_j)$. In general, it has been established that in order for a hash function $h$ to have the property that $P[h(B_i) = h(B_j)] = J_B(B_i, B_j)$, it must be an instance of Consistent Sampling~\citep{manesse_cws2010} (see Definition~\ref{df:cs}).

\begin{df}[Consistent Sampling~\citep{manesse_cws2010}]
  \label{df:cs}
  Given a bag $B_i$ with multiplicities $B_i^w, w = 1, \ldots, D$, consistent sampling generates a sample $(w,z_w): 0 \leq z_w \leq B_i^w$ with the following two properties.
  \end{df}
  \begin{enumerate}
  \item \emph{Uniformity}: Each sample $(w, z_w)$ should be drawn uniformly at random from $\bigcup\limits_{w=1}^D \{\{w\} \times [0, B_i^w]\}$, where $B_i^w$ is the multiplicity of the element $w$ in $B_i$. In other words, the probability of drawing $w$ as a sample of $B_i$ is proportional to its multiplicity $B_i^w$ and $z_w$ is uniformly distributed.
  \item \emph{Consistency}: If $B_j^w \leq B_i^w, \forall w$, then any sample $(w, z_w)$ drawn from $B_i$ that satisfies $z_w \leq B_j^w$ will also be a sample from $B_j$.
  \end{enumerate}

Once the $M$ MinHash values for each bag have been computed, $l$ tuples $g_1, \ldots, g_l$ of $r$ different MinHash values are defined as follows

\begin{equation*}
  \begin{tabular}{l}
    $g_1(B_i) = (h_1(B_i), h_2(B_i), \ldots , h_r(B_i))$\\
    $g_2(B_i) = (h_{r+1}(B_i), h_{r+2}(B_i), \ldots , h_{2\cdot r}(B_i))$\\
    $\cdots$ \\
    $g_l(B_i) = (h_{(l-1)\cdot r+1}(B_i), h_{(l-1)\cdot r+2}(B_i), \ldots , h_{l\cdot r}(B_i))$
  \end{tabular},
\end{equation*}

\noindent where $h_m(B_i), m \in \{1, \ldots, r \cdot l\}$ is the $m$-th MinHash value of bag $B_i$ and $M = r \cdot l$.
Thus, $l$ different hash tables are constructed and each bag $B_i$ is stored in the bucket corresponding to $g_x(B_i)$ for each hash table $x = 1, \ldots, l$.
Two bags $\{B_i, B_j\}$ are stored in the same hash bucket on the $x$-th hash table iff 
$g_x(B_i) = g_x(B_j), x \in \{1, \ldots, l\}$, i.e. all the MinHash values of the tuple $g_x$ are the same for both bags.
Since similar bags are expected to share several MinHash values, there is a high probability that they will have an identical tuple. 
In contrast, dissimilar bags will seldom have the same MinHash value, and therefore the 
probability that they will have an identical tuple will be low. More precisely, the probability 
that two bags $B_i$ and $B_j$ share the $r$ different MinHash values of a given tuple $g_x, x \in \{1, \ldots l\}$ is

\begin{equation}
  P[g_x(B_i) = g_x(B_j)] = J_B(B_i,B_j)^r.
\end{equation}

\noindent Consequently, the probability that two bags $B_i$ and $B_j$ have at least one identical tuple is

\begin{equation}
  P_{collision}[B_i,B_j] = 1-(1-J_B(B_i,B_j)^r)^l.
\end{equation}

To search for similar bags to a given query bag $Q$, first the $l$ different tuples $g_1(Q), \ldots, g_l(Q)$ are computed for Q. 
Then, the corresponding buckets in the $l$ hash tables are inspected and all stored bags $B_i$ for which $g_x(B_i) = g_x(Q), x = 1, \ldots, l$ are retrieved.
Finally, the retrieved bags are sorted in descending order of their Jaccard similarity $J_B(Q, B_i)$ with the query bag $Q$;
typically, retrieved bags with lower similarity than a given threshold are discarded. 

\section{Sampled Min-Hashing for Topic Discovery}
\label{sec:smh}
\subsection{Min-Hashing for Mining Beyond-Pairwise Relationships}
In order to measure beyond-pairwise relationships between multiple sets, the Jaccard similarity in Eq.~\ref{eq:mh_prob} can be generalized as the \emph{Jaccard Co-Occurrence Coefficient} for $k$ sets $\{S^{(1)}, \ldots , S^{(k)}\} \subseteq \{S_1, \ldots , S_N\}, k \in 2, \ldots , N$ as follows

\begin{equation}
  \label{eq:jcc}
  JCC(S^{(1)}, \ldots , S^{(k)}) = \frac{\vert S^{(1)} \cap \cdots \cap S^{(k)} \vert }{\vert S^{(1)} \cup \cdots \cup S^{(k)} \vert},
\end{equation}

\noindent where the numerator is the number of elements that all the sets $\{S^{(1)}, \ldots , S^{(k)}\}$ have in common,
and the denominator corresponds to the number of elements that appear at least once in the sets $\{S^{(1)}, \ldots , S^{(k)}\}$.

The property that a MinHash $h$ adheres to Eq.~\ref{eq:mh_prob} can be directly extended to $k$ sets~\citep{fuentes, shrivastava2013beyond}, i.e.

\begin{equation}
  \label{eq:jcc_prob}
  P[h(S^{(1)}) = \ldots = h(S^{(k)})] = JCC(S^{(1)}, \ldots , S^{(k)}).
\end{equation}

More generally, we can define the \emph{Jaccard Co-Occurrence Coefficient} for $k$ bags $\{B^{(1)}, \ldots , B^{(k)}\} \subseteq \{B_1, \ldots , B_N\}, k \in 2, \ldots, N$ as 

\begin{equation}
    \label{eq:jccb}
  \mathit{JCC}_B(B^{(1)}, \ldots , B^{(k)}) = \frac{\sum_w \min{(B^{(1)^w}, \ldots , B^{(k)^w})}}{\sum_w \max{(B^{(1)^w}, \ldots , B^{(k)^w})}} \in [0,1], 
\end{equation}

\noindent where $B^{(1)^w}, \ldots, B^{(k)^w}$ are the multiplicities of the element $w$ in bags $B^{(1)}, \ldots, B^{(k)}$~respectively. From Definition~\ref{df:cs}, it follows that

\begin{equation}
  \label{eq:jccb_prob}
  P[h(B^{(1)}) = \ldots = h(B^{(k)})] = JCC_B(B^{(1)}, \ldots , B^{(k)}),
\end{equation}

\noindent for any hash function $h$ generated with consistent sampling. Eq~\ref{eq:jccb_prob} holds because the $k$ bags $\{B^{(1)^w}, \ldots , B^{(k)^w}\}$ will have an identical MinHash value every time the sample $(w, z_k)$ from the maximum multiplicity $\max{(B^{(1)^w}, \ldots , B^{(k)^w})}$ is less than or equal to the minimum multiplicity $\min{(B^{(1)^w}, \ldots , B^{(k)^w})}$, which in general will happen $\frac{\sum_w \min{(B^{(1)^w}, \ldots , B^{(k)^w})}}{\sum_w \max{(B^{(1)^w}, \ldots , B^{(k)^w})}}$ times given that all samples are drawn uniformly at random from the multiplicity of each bag.

As in Min-Hashing for pairwise similarity search, $l$ tuples of $r$ MinHash values are computed and the probability that $k$ bags will have an identical tuple $g_x$ is given by
\begin{equation}
  P[g_x(B^{(1)}) =  \cdots = g_x(B^{(k)})] = JCC_B(B^{(1)},  \ldots , B^{(k)})^r.
\end{equation}

\begin{figure}[tb]
  \centering
  \includegraphics[width=\textwidth]{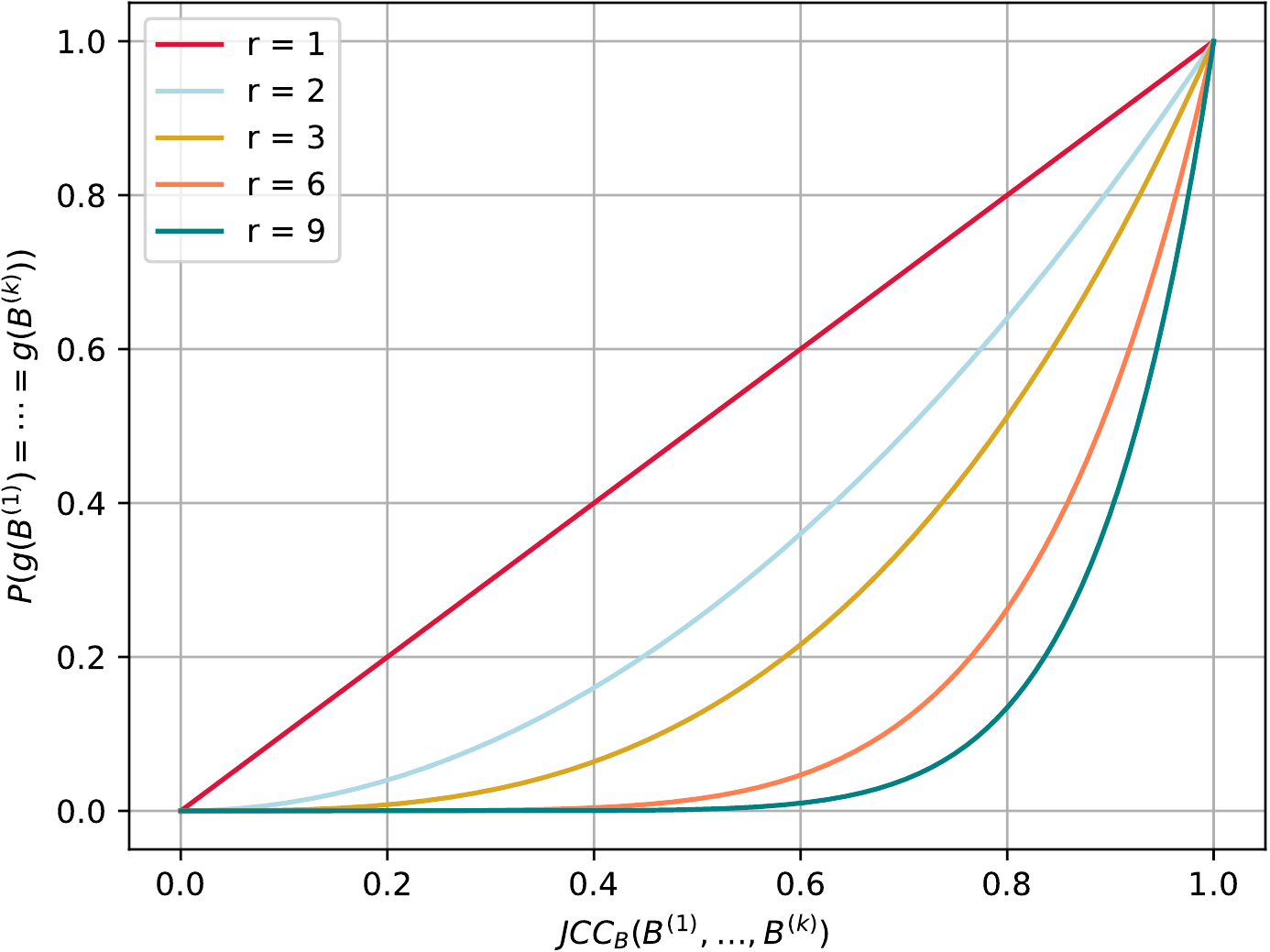}
  \caption{Probability of bags $\{B^{(1)}, \ldots , B^{(k)}\}$ having an identical tuple as a function of their $JCC_B$ for different tuple sizes ($r$).}
  \label{fig:jcc_collision}
\end{figure}

Figure.~\ref{fig:jcc_collision} shows the plots of the probability of $k$ bags having an identical tuple as a function of their $JCC_B$ for different tuple sizes $r$.
As can be observed, the probability increases with larger $JCC_B$ values while it decreases exponentially for larger tuple sizes $r$.
Having larger tuple sizes allows us to reduce the probability that bags with small $JCC_B$ values have an identical tuple, but comes with the cost of also reducing the probability that bags with larger $JCC_B$ values have an identical tuple.
However, we can increase the latter probability by increasing the number of tuples $l$.
Specifically, the probability that $k$ bags $\{B^{(1)}, \ldots , B^{(k)}\}$ have at least one identical tuple from the $l$ different tuples is

\begin{equation}
  P_{collision}[B^{(1)}, \ldots , B^{(k)}] = 1-(1-JCC_B(B^{(1)},  \ldots , B^{(k)})^r)^l.
\end{equation}

\noindent Therefore, the choice of $r$ and $l$ becomes a trade-off between precision and recall.

\begin{figure}[t]
  \centering
  \includegraphics[width=\textwidth]{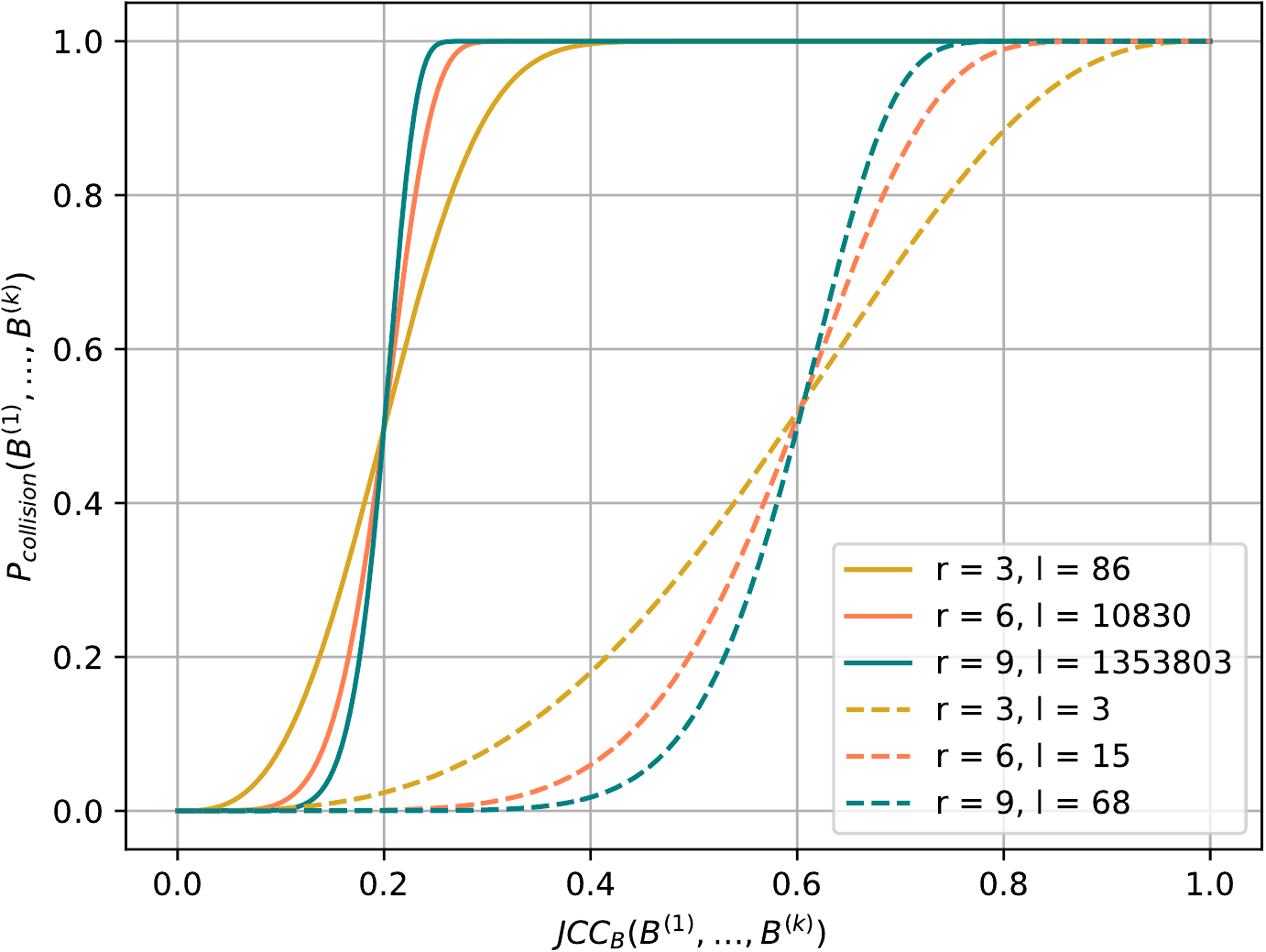}
  \caption{Collision probability of bags $\{B^{(1)}, \ldots , B^{(k)}\}$ as a function of their $JCC_B$ for different co-occurrence thresholds ($\eta$) and tuple sizes ($r$).}
  \label{fig:jcc_unit}
\end{figure}

As illustrated in Fig.~\ref{fig:jcc_unit}, the probability that $k$ bags have at least one identical tuple approximates a co-occurrence filter such that

\begin{equation}
  \label{eq:jcc_unit}
  P_{collision}[B^{(1)}, \ldots , B^{(k)}] \approx
  \begin{cases}
    1 & \mbox{if } JCC_B(B^{(1)}, \ldots , B^{(k)}) \geq \eta \\
    0 & \mbox{if } JCC_B(B^{(1)}, \ldots , B^{(k)}) < \eta
  \end{cases},
\end{equation}

\noindent where $\eta$ is a $JCC_B$ threshold parameter of the filter, defined by the user. Given the $JCC_B$ threshold $\eta$ and the tuple size $r$, we can obtain the number of tuples $l$ by setting $P_{collision}$ to $0.5$ and solving for $l$, which gives

\begin{equation}
\label{eq:s}
  l = \frac{\log(0.5)}{\log(1 - \eta^r)}.
\end{equation}

\noindent Note that the number of tuples $l$ increases exponentially as the tuple size $r$ increases and/or the $JCC_B$ threshold $\eta$ decreases.

\subsection{Topic Discovery}

Finding word co-occurrences has been a recurrent task in Natural Language Processing for several decades as they underlie different linguistic phenomena such as semantic relationships or lexico-syntactic constraints.
Here we hypothesize that highly co-occurring words likely belong to the same topic, and propose to mine such words by applying Min-Hashing to the occurrence pattern of each word in a given corpus. 
In this approach, the degree of co-occurrence of a set of words is measured by their $JCC_B$ (see Eq.~\ref{eq:jccb}), which corresponds to the number of documents where all the words occur together divided by the number of documents where at least one of the words occur.
Although other measures of beyond-pairwise co-occurrence could be employed, we rely on $JCC_B$ because sets of words with high $JCC_B$ can be efficiently found using Min-Hashing.

To discover topics we represent each document in the corpus by a bag-of-words $B_i, i = 1, \ldots , N$ and the occurrence pattern of each word $w_j, j = 1, \ldots, D$ in the vocabulary by its corresponding inverted file bag $B_j', j = 1, \ldots , D$ whose elements are document IDs and whose multiplicities $B'^{s}_j, s = 1, \ldots , N$ are the frequencies with which the word $w_j$ occurred in the document $s$.
After computing $l$ tuples and storing each inverted file bag $B'_1, \ldots , B'_D$ in the corresponding $l$ hash tables, we extract each set $C_y, y = 1, \ldots, Y$ composed of $k$ inverted file bags $\{B'^{(1)}, \ldots , B'^{(k)}\}$ with an identical tuple $g_x(B'^{(1)}) = \cdots = g_x(B'^{(k)}), x \in \{1, \ldots, l\}$ (i.e. they are stored in the same bucket in the same hash table), where $k \geq 3$ since we are considering beyond-pairwise co-occurrences.
We call these sets \emph{co-occurring word sets (CWS)} because they are composed of inverted file bags corresponding to words with high $JCC_B$ values.
The Min-Hashing parameter $\eta$ (see Eq.~\ref{eq:jcc_unit}) controls the degree of co-occurrence of the words in each CWS so that higher values of $\eta$ will produce CWS with higher $JCC_B$ values whereas smaller values of $\eta$ will produce CWS with smaller $JCC_B$ values.
In order to reduce the space complexity, since the $l$ tuples are generated from independent hash functions, we can compute them one by one so that only one hash table (instead of $l$) is maintained in memory at every moment.

We name this approach \emph{Sampled Min-Hashing (SMH)} because each hash table associated to a tuple generates CWS by sampling the word occurrence space spanned by the inverted file bags $\{B'_1, \ldots , B'_D\}$, that is, each hash table randomly partitions the vocabulary based on the word occurrences.
In SMH, multiple random partitions are induced by different hash tables, each of which generates several CWS.
Representative and stable words belonging to the same topic are expected to be present in multiple CWS (i.e. lie on overlapping inter-partition cells).
Therefore, we cluster CWS that share many words in an agglomerative manner to form the final topics $T_a, a = 1, \ldots, A$.
We measure the proportion of words shared between two CWS $C_i$ and $C_j$ by their overlap coefficient, namely

\begin{equation*}
  ovr(C_i,C_j) = \frac{\mid C_i \cap C_j \mid}{\min(\mid C_i \mid,\mid C_j\mid )} \in [0,1].
\end{equation*}

\begin{figure}[tb]
  \centering
  \includegraphics[width=\textwidth]{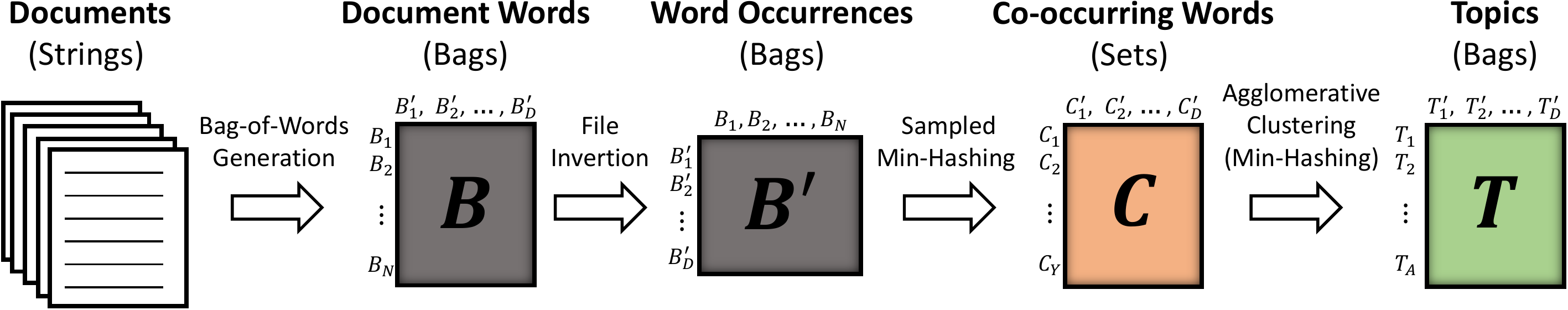}
  \caption{Overview of topic discovery by Sampled Min-Hashing.}
  \label{fig:overview}
\end{figure}

\noindent This agglomerative clustering can be formulated as finding the connected components of an undirected graph whose vertices are the CWS and edges connect every pair with an overlap coefficient greater than a threshold $\epsilon$, i.e. $G = (\{C_1, \ldots,  C_Y\}, \{(C_i, C_j):ovr(C_i, C_j) > \epsilon, i, j = 1, \ldots, Y, i \neq j\})$.
Each connected component of $G$ is a cluster composed of the CWS that form a topic.
Given that $J(C_i, C_j) \leq ovr(C_i, C_j),  \forall i,j$, we can efficiently find these CWS pairs by using Min-Hashing for pairwise similarity search (see Sect.~\ref{sec:minhash}), thus avoiding the overhead of computing the overlap coefficient between all pairs.
An overview of the whole topic discovery process by SMH can be seen in Fig.~\ref{fig:overview}.

The agglomerative clustering merges chains of CWS with high overlap coefficients into the same topic.
As a result, CWS associated with the same topic can belong to the same cluster even if they do not share words with one another, as long as they are members of the same chain.
In general, the generated clusters have the property that for any CWS, there exists at least one CWS in the same cluster with which it has an overlap coefficient greater than a given threshold $\epsilon$.
Note that this is a connectivity-based clustering procedure which generates clusters based on the minimum similarity of all pairs of sets.
Because of this, the number of topics $A$ produced by SMH depends on the choice of parameter values and word co-occurrence characteristics of the corpus.
This contrasts with LDA and other topic models where the number of topics is given in advance by the user.

Finally, each topic $T_a, a = 1, \ldots, A$ discovered by SMH is represented by the set of all words in the CWS that belong to the topic.
Therefore, the number of words in a topic also depends on the parameter configuration and the degree of co-occurrence of words belonging to the same topic in the corpus. 
This again contrasts with LDA and other topic models where topics are represented as distributions over the complete vocabulary, although only the top-$K$ most probable words (typically $K$ is set to 10) of each topic are shown to the user.
For each topic, words are sorted in descending order of the number of CWS in which they appear, such that more representative and coherent words are shown first to the user.

\section{Experimental Results}
\label{sec:eval}
We evaluated~\footnote{The source code for all the reported experiments related to topic discovery is available at~\url{https://github.com/gibranfp/SMH-Topic-Discovery}. An implementation of Sampled Min-Hashing is available at~\url{https://github.com/gibranfp/Sampled-MinHashing}.} the coherence of the topics discovered by SMH on the Reuters corpus~\citep{rose2002reuters} using different parameter settings and vocabulary sizes.
We also compared SMH to Online LDA with respect to both topic coherence and scalability using corpora of increasing sizes.
Specifically, we performed experiments on 20 Newsgroups (a collection of $18,846$ newsgroup documents), Reuters (a collection of $806,791$ news articles), and Spanish and English Wikipedia (2 collections of $1,286,095$ and $5,228,998$ encyclopedia entries, respectively)\footnote{English Wikipedia dump from 2016--11--01. Spanish Wikipedia dump from 2017--04--20.}.
In all 4 corpora, a standard list of top words were removed and the remaining vocabulary was restricted to the $D$ most frequent words ($D = 20,000$ for 20 Newsgroups, $D = 100,000$ for Reuters and $D = 1,000,000$ for both Spanish and English Wikipedia). 
It is worth noting that these vocabulary sizes are considerably larger than what it is typically used in topic models (e.g. in~\citep{onlinelda} topics were discovered from $352,549$ Nature articles using a vocabulary of $4,253$ words and from $100,000$ Wikipedia articles using a vocabulary of $7,995$ words).
Here, we decided to use larger vocabulary sizes in order to evaluate the scalability and robustness of SMH with respect to both corpus and vocabulary size.

\begin{figure}[!tb]
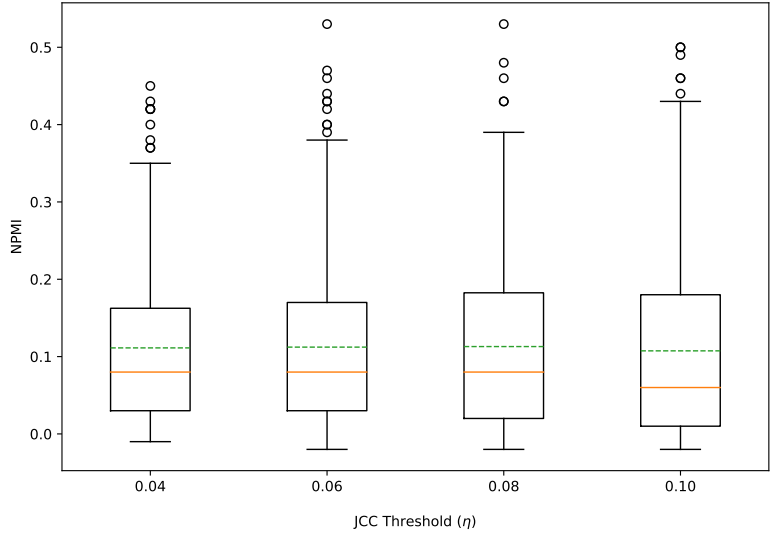

  \centering
  \includegraphics[width=\textwidth]{{{coocurrence_threshold.k400}}}
  \caption{NPMI scores for topics discovered by SMH with different $JCC_B$ thresholds ($\eta$). Median NPMI is shown as a solid yellow line and mean NPMI as a dashed green line.}
  \label{fig:cothres}
\end{figure}

In order to evaluate topic coherence, we relied on the Normalized Point Mutual Information (NPMI) since it strongly correlates with human judgments and outperforms other metrics~\citep{npmi}.
NPMI is defined for an ordered topic $T$ from its top-$K$ words as follows

\begin{equation}
\begin{split}
  NPMI(T)=\sum_{j=2}^K \sum_{i=1}^{j-1} \frac{ \log{\frac{P(w_j,w_i)}{P(w_i)P(w_j)}}}{-\log {P(w_i,w_j)}}
\end{split}
\end{equation}

\noindent Following the evaluation methodology of~\citet{npmi}~\footnote{We used the implementation provided by the authors, which is available at \url{https://github.com/jhlau/topic_interpretability}}, all 4 corpora were lemmatized using NLTK's WordNet lemmatizer~\citep{nltk}.
NPMI scores were then computed from the top-10 words for each topic, and lexical probabilities $P(w_i, w_j)$, $P(w_i)$ and $P(w_j)$ were calculated by sampling word counts within a sliding context window over an external corpus, in this case the lemmatized English Wikipedia. 

As mentioned in Sect.~\ref{sec:smh}, in SMH both the number of topics and the number of words in each topic depend on the parameter setting and the corpus characteristics.
In order to make the evaluation of all models comparable, we sorted topics in descending order of the average number of documents in which their top-10 words appear (all topics with less than 10 words were discarded), and only took into account the top 400 topics.
In addition, only clusters with at least 5 CWS were considered in order to avoid random topics that may not be meaningful.

\begin{table}[tb]
\centering
\caption{NPMI statistics for SMH with different $JCC_B$ thresholds ($\eta$).}
\label{table:jcc}
\begin{tabular}{cccccccccc}
\toprule
 & & \multicolumn{3}{c}{\textbf{NPMI}} & & & &  \\
 \cmidrule{3-5} 
 \multicolumn{1}{c}{\textbf{$\eta$}} & & \textbf{Avg} & \textbf{Med} & \textbf{STD} & & \textbf{\#Topics} & & \textbf{Time}\\
 \cmidrule{1-1} \cmidrule{3-5}  \cmidrule{7-7} \cmidrule{9-9}
0.04 & & 0.111 & 0.08 & 0.100 & & 1708 & & 2111\\
0.06 & & 0.112 & 0.08 & 0.106 & & 1038 & & 1288\\
0.08 & & 0.112 & 0.08 & 0.108 & &  743 & & 591\\
0.10 & & 0.107 & 0.06 & 0.115 & &  572 & & 358\\
\bottomrule
\end{tabular}
\end{table}

\subsection{Evaluation of SMH Parameters}
SMH has 3 main parameters that can affect its behavior and output: the $JCC_B$ threshold $\eta$, the tuple size $r$, and the overlap coefficient $\epsilon$.
We ran experiments on the Reuters corpus using a range of different values for these parameters in order to evaluate their impact on the time required to discover topics, the number of discovered topics, and the coherence of the top 400 topics.
In the following, we describe each of these experiments in detail and discuss the results.

The $JCC_B$ threshold $\eta$ is an SMH parameter that roughly controls to what degree a group of words must co-occur in order to be stored in the same bucket in at least one hash table (see Eq.~\ref{eq:jcc_unit} and Fig.~\ref{fig:jcc_unit}) and therefore be considered a co-occurring word set (CWS).
For large $\eta$ values, words need to have a higher co-occurrence (i.e. have a higher $JCC_B$) to be considered a CWS.
Conversely, smaller $\eta$ values are more permissive and allow words with lower co-occurrence to be considered a CWS.
Consequently, smaller $\eta$ values require more tuples than larger $\eta$ values.
We evaluate the coherence of the topics discovered by SMH using $\eta$ values of $0.04$, $0.06$, $0.08$, and $0.10$.
By setting the tuple size $r$ to 2 and using Eq.~\ref{eq:s}, we found the number of tuples (hash tables) $l$ for these $\eta$ values to be $432$, $192$, $107$ and $68$, respectively.
Figure \ref{fig:cothres} shows the distribution of NPMI scores for the 4 different $\eta$ values.
Interestingly, the coherence of the discovered topics remains stable over the range of $\eta$ values and only noticeably declines when $\eta = 0.10$.
As shown in Table \ref{table:jcc}, the $\eta$ value has a greater impact on the number of discovered topics, reducing quickly as $\eta$ increases.
This is because many relevant CWS tend to lie towards small $JCC_B$ values and are therefore not found by SMH with larger $\eta$ values.
We can also observe that the discovery time grows rapidly as $\eta$ decreases, since smaller $\eta$ values require more tuples to be computed.
So, smaller $\eta$ values may improve recall but at the cost of increasing discovery time.

\begin{table}[tb]
\centering
\caption{NPMI statistics for SMH using different tuple sizes ($r$).}
\label{table:tupsize}
\begin{tabular}{ccccccccc}
\toprule
& & \multicolumn{3}{c}{\textbf{NPMI}} & & \\
\cmidrule{3-5} 
$r$ & & \textbf{Avg} & \textbf{Med} & \textbf{STD} & & \textbf{\#Topics} & & \textbf{Time}\\
\cmidrule{1-1} \cmidrule{3-5}  \cmidrule{7-7} \cmidrule{9-9}
2 & & 0.112 & 0.08 & 0.108 & & 743 & & 743\\
3 & & 0.115 & 0.06 & 0.119 & & 580 & & 9588\\
4 & & 0.120 & 0.07 & 0.122 & & 544 & & 139954\\
\bottomrule
\end{tabular}
\end{table}

The tuple size $r$ determines how closely the probability of finding a CWS approximates a unit step function such that only CWS with a $JCC_B$ larger than $\eta$ are likely found by SMH.
We evaluate SMH with different tuple sizes, specifically $r$ equal to $2$, $3$ and $4$.
Again, by using Eq.~\ref{eq:s} we found that the number of tuples $l$ for these tuple sizes and $\eta = 0.08$ is $107$, $1353$ and $16922$, respectively.
Table~\ref{table:tupsize} shows the NPMI statistics, the number of discovered topics, and discovery time for all 3 different tuple sizes. 
Note that the average and median NPMI as well as the standard deviation are very similar for the 3 tuple sizes.
On the other hand, the number of discovered topics consistently decreases for larger tuple sizes, since the probability of finding a CWS more closely approximates a unit step function and as a result there are less false positives.
However, the discovery time grows exponentially with the tuple size because a significantly larger number of tuples are required.
Therefore, a larger tuple size $r$ may improve precision but at a high computational cost.

Finally, we evaluate the impact of the overlap coefficient $\epsilon$.
This parameter specifies the degree of overlap that 2 CWS must have in order to be merged into the same cluster and become the same topic.
Small $\epsilon$ values allow pairs of CWS that have a small proportion of shared words to be merged into the same cluster.
In contrast, larger $\epsilon$ values require a larger proportion of shared words.
Table \ref{table:overlap} presents NPMI statistics as well as the number of discovered topics and the discovery time for SMH with different $\epsilon$ values.
We can observe that NPMI scores were considerably lower for $\epsilon = 0.5$ than for $\epsilon = 0.7$ and $\epsilon = 0.9$, while the number of discovered topics and the discovery speed was very similar for the 3 $\epsilon$ values.
The reason $\epsilon = 0.5$ produces topics with lower NPMI scores is that the threshold becomes too low, which causes many CWS from different topics to be merged into a single topic.

\begin{table}[tb]
\centering
\caption{NPMI statistics for SMH with different overlap coefficient thresholds ($\epsilon$).}
\label{table:overlap}
\begin{tabular}{ccccccccc}
\toprule
& & \multicolumn{3}{c}{\textbf{NPMI}} & & & & \\
\cmidrule{3-5} 
$\epsilon$ & & \textbf{Avg} & \textbf{Med} & \textbf{STD} & & \textbf{\#Topics} & & \textbf{Time}\\
\cmidrule{1-1} \cmidrule{3-5} \cmidrule{7-7} \cmidrule{9-9}
0.5 & & 0.063 & 0.04 & 0.066 & & 1842 & & 2278\\
0.7 & & 0.101 & 0.07 & 0.093 & & 1795 & & 2066\\
0.9 & & 0.111 & 0.08 & 0.100 & & 1708 & & 2111\\
\bottomrule
\end{tabular}
\end{table}

\subsection{Impact of the Vocabulary Size}
Reducing the vocabulary to the top-$D$ most frequent words is a common approach to improve the quality of the discovered topics and speed up discovery.
Here, we evaluate the impact of different vocabulary sizes on the coherence of the discovered topics by SMH with $r = 2$, $\eta = 0.04$ and $\epsilon = 0.9$.
Table~\ref{table:vocsize} shows NPMI statistics, the number of discovered topics and discovery time for the Reuters corpus with vocabularies composed of the top $20,000$, $40,000$, $60,000$, $80,000$ and $100,000$ words.
In general, NPMI scores decrease as the vocabulary size increases.
This is expected since larger vocabularies introduce less common words which may not appear frequently in the reference corpus from which NPMI's lexical probabilities are sampled.
However, the number of discovered topics consistently increases with larger vocabularies because additional topics are formed with the extra words.
Surprisingly, the discovery time was very similar for the 5 different vocabulary sizes despite there being 5 times more words in the largest vocabulary than there were in the smallest. 

\begin{table}[tb]
\centering
\caption{NPMI statistics for SMH with different vocabulary sizes ($D$).}
\label{table:vocsize}
\begin{tabular}{ccccccccc}
\toprule
& & \multicolumn{3}{c}{\textbf{NPMI}} & & \\
\cmidrule{3-5}
$D$ & & \textbf{Avg} & \textbf{Med} & \textbf{STD} & & \textbf{\#Topics} & & \textbf{Time}\\
\cmidrule{1-1} \cmidrule{3-5}  \cmidrule{7-7} \cmidrule{9-9}
20000 & & 0.130 & 0.11 & 0.109 & & 334 & & 2120\\
40000 & & 0.121 & 0.10 & 0.102 & & 610 & & 2162\\
60000 & & 0.114 & 0.09 & 0.098 & & 909 & & 2044\\
80000 & & 0.115 & 0.09 & 0.100 & & 1228 & & 2072\\
100000 & & 0.111 & 0.08 & 0.101 & & 1708 & & 2111\\
\bottomrule
\end{tabular}
\end{table}

\subsection{Comparison with Online LDA}
\begin{figure}[!tb]
  \centering
  \includegraphics[width=\textwidth]{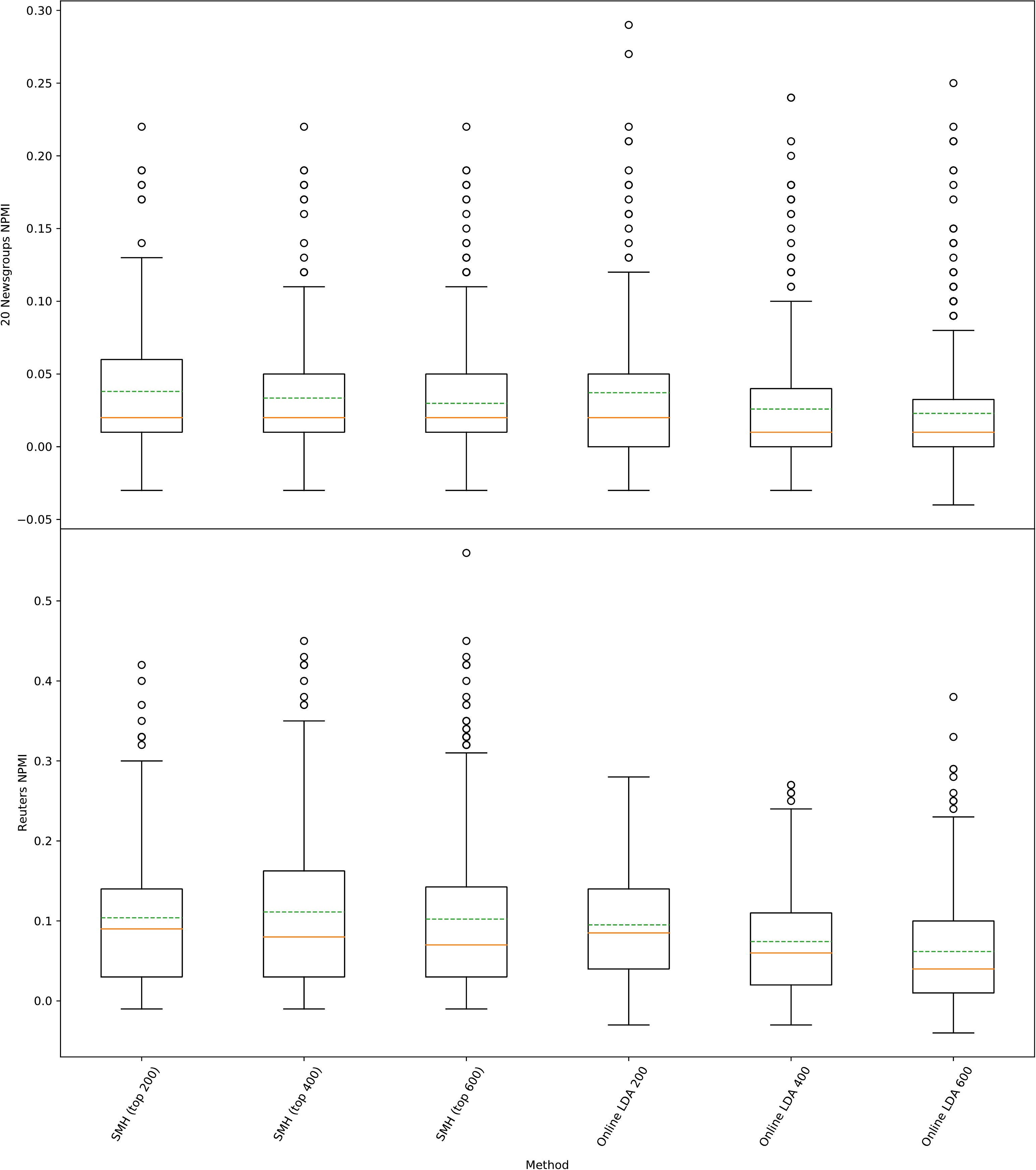}
  \caption{NPMI scores for SMH (top 200, 400 and 600) and Online LDA (topic number set to 200, 400 and 600) topics discovered from 20 Newsgroups (top) and Reuters (bottom). Median NPMI is shown as a solid yellow line and mean NPMI as a dashed green line.}
  \label{fig:comparison}
\end{figure}

LDA has been the dominant approach to topic discovery for over a decade.
Therefore, we compare the coherence of SMH and Online LDA topics using the 20 Newsgroups and Reuters corpora.
Online LDA uses stochastic variational inference to approximate the posterior distribution that defines the topics of the corpus\footnote{We used the implementation included in scikit-learn~\citep{sklearn}, which is based on the code originally provided by the authors. The learning parameters were set to the values with the best performance reported by~\citet{onlinelda}.}.
It allows for topic discovery at a larger scale without the need of a computer cluster and has become a popular alternative to sampling approaches.
The NPMI scores for topics discovered by SMH with $r = 2$, $\eta = 0.04$ and $\epsilon = 0.9$ (top 200, 400 and 600 topics) and Online LDA (number of topics set to 200, 400 and 600) are shown in Figure~\ref{fig:comparison}.
For both corpora, the distributions of NPMI scores of SMH and Online LDA topics are very similar.
Note that increasing the number of topics tends to shift the distribution of NPMI scores for both approaches towards lower values, since more topics with less common words are considered. However, the effect is more severe in Online LDA than SMH.

\subsection{Scalability}
In order to evaluate the scalability of SMH, we discovered topics from the 20 Newsgroups, Reuters, and Spanish and English Wikipedia corpora whose sizes range from thousands to millions of documents and whose vocabularies range from thousands of words to as much as one million words.
We also compared the time required by SMH to discover topics with Online LDA\footnote{Due to the high memory and computational requirements, it was not possible to run Online LDA for the Spanish and English Wikipedia.}. 
All experiments were performed on a Dell\textsuperscript{TM} PowerEdge\textsuperscript{TM} with 2 Intel\textsuperscript{\textregistered} Xeon\textsuperscript{\textregistered} CPUs X5650\at 2.67GHz (12 cores) and 32 GB of RAM.
For comparison purposes, each experiment used only a single thread. 
Table~\ref{table:times} presents the discovery time in seconds for SMH with tuple size $r = 2$ and $JCC_B$ thresholds $\eta = \{0.04, 0.06, 0.08, 0.10\}$, compared with Online LDA at $200$, $400$ and $600$ topics.
Note that the time complexity of SMH is linear with respect to both corpus and vocabulary size.
Remarkably, SMH took at most 6.4 hours (when $\eta = 0.04$) and as little as 58 minutes (when $\eta = 0.10$) to process the entire Wikipedia in English, which contains over 5 million documents with a vocabulary of 1 million words.
Although the time required by both SMH and Online LDA to process the 20 Newsgroups corpus was very similar, for the Reuters corpus SMH was significantly faster than Online LDA.

\begin{table}[tb]
\centering
\caption{Time in seconds to discover topics on the 20 Newsgroups, Reuters and Wikipedia corpora with vocabularies of 10000, 100000 and 1000000 words, respectively.}
\label{table:times}
\begin{tabular}{lccccccccc}
\toprule
& & \multicolumn{4}{c}{\textbf{SMH}} & & \multicolumn{3}{c}{\textbf{Online LDA}} \\
\cmidrule{3-6} \cmidrule{8-10}
\textbf{Corpus} & & \textbf{0.04} & \textbf{0.06} & \textbf{0.08} & \textbf{0.10} & & \textbf{200} &  \textbf{400} & \textbf{600}\\
\cmidrule{1-1} \cmidrule{3-6} \cmidrule{8-10}
20 Newsgroups & & 158 & 49 & 26 & 16 & & 138 & 236 & 311 \\
Reuters & & 2111 & 1288 & 591 & 358 & & 9744 & 101418 & 138144\\
Wikipedia (Es) & & 8173 & 4181 & 2332 & 1483 & & -- & -- & --\\
Wikipedia (En) & & 22777 & 10669 & 5353 & 3475 & & -- & -- & --\\
\bottomrule
\end{tabular}
\end{table}

\subsection{Examples of Discovered Topics by SMH}
\begin{table}[tbhp]
  \caption{Sample topics discovered by SMH from the 20 Newsgroups, Reuters, and Spanish and English Wikipedia corpora with vocabulary size ($D$) of 20K, 100K, 1M, and 1M words, respectively. For each topic, its size and the top-10 words are presented.}
  \scriptsize
  \centering
  \begin{tabular}{|c|l|}\hline
  {\centering \bf Size} & \multicolumn{1}{c|}{\centering \bf Top 10 words} \\\hline
\multicolumn{2}{|c|}{\bf 20 Newsgroups ($D=20K$)}\\\hline
$36$ & religion, atheist, religious, atheism, belief, christian, faith, argument, bear, catholic,\ldots\\\hline
$13$ & os, cpu, pc, memory, windows, microsoft, price, fast, late, manager,\ldots \\\hline
$31$ & game, season, team, play, score, minnesota, win, move, league, playoff,\ldots \\\hline
$23$ & rfc, crypt, cryptography, hash, snefru, verification, communication, privacy, answers, signature,\ldots  \\\hline
$45$ & decision, president, department, justice, attorney, question, official, responsibility, yesterday, conversation,\ldots \\\hline
$19$ & meter, uars, balloon, ozone, scientific, foot, flight, facility, experiment, atmosphere,\ldots \\\hline
$61$ & dementia, predisposition, huntington, incurable, ross, forgetfulness, suzanne, alzheimer, worsen, parkinson,\ldots \\\hline
\multicolumn{2}{|c|}{\bf Reuters ($D=100K$)}\\\hline
$93$ & point, index, market, high, stock, close, end, share, trade, rise,\ldots \\\hline
$85$ &  voter, election, poll, party, opinion, prime, seat, candidate, presidential, hold,\ldots\\\hline
$79$ & play, team, match, game, win, season, cup, couch, final, champion,\ldots \\\hline
$68$ & wrongful, fujisaki, nicole, acquit, ronald, jury, juror, hiroshi, murder, petrocelli,\ldots \\\hline
 $12$ & window, nt, microsoft, computer, server, software, unix, company, announce, machine,\ldots\\\hline
$28$ & mexico, mexican, peso, city, state, trade, foreign, year, share, government,\ldots  \\\hline
$87$ & spongiform, encephalopathy, bovine, jakob, creutzfeldt, mad, cow, wasting, cjd, bse,\ldots \\\hline
 \multicolumn{2}{|c|}{\bf Wikipedia Spanish ($D=1M$)}\\\hline
 $179$ & amerindios, residiendo, afroamericanos, hispanos, isleños, asiáticos, latinos, pertenecían, firme, habkm²,\ldots $ $\\\hline
 $32$ & river, plate, juniors, boca, racing, libertadores, clubes, posiciones, lorenzo, rival,\ldots \\\hline
 $162$ & depa, billaba, obiwan, kenobi, padmé, haruun, vaapad, amidala, syndulla, skywalker,\ldots $$\\\hline 
 $28$ & touchdowns, touchdown, quarterback, intercepciones, pases, yardas, nfl, patriots, recepciones, jets,\ldots $ $\\\hline
$58$ & canción, disco, álbum, canciones, sencillo, unidos, reino, unido, número, música,\ldots $ $\\\hline
$69$ & poeta, poesía, poemas, poetas, mundo, escribió, literatura, poema, nacional, siglo,\ldots \\\hline
 \multicolumn{2}{|c|}{\bf Wikipedia English ($D=1M$)}\\\hline
 $198$ & families, householder, capita, makeup, latino, median, hispanic, racial, household, census,\ldots \\\hline
 $30$ & dortmund, borussia, schalke, leverkusen, werder, bayer, eintracht, wolfsburg, vfl, vfb,\ldots \\\hline 
 $209$ & padmé, luminara, amidala, barriss, talzin, offee, unduli, skywalker, darth, palpatine,\ldots \\\hline
 $97$ & touchdown, yard, quarterback, pas, quarter, interception, fumble, rush, sack, bowl,\ldots \\\hline 
 $163$ & release, album, guitar, single, bass, vocal, band, drum, chart, record,\ldots \\\hline
 $67$ & vowel, consonant, noun, plural, verb, pronoun, syllable, tense, adjective, singular,\ldots \\\hline
$14$ & mesoamerican, mesoamerica, olmec, michoacán, preclassic, abaj, takalik, exact, corn, veracruz,\ldots \\\hline
  \end{tabular}
  \label{tbl:examples}
\end{table}

 Table~\ref{tbl:examples} exemplifies some of the topics discovered by SMH on the 20 Newsgroups, Reuters, and English and Spanish Wikipedia corpora\footnote{The complete set of topics discovered by SMH on each corpus is available at \url{https://github.com/gibranfp/SMH-Topic-Discovery/blob/master/example_topics/}}. 
In general, these topics range from small (tens of words) to large (hundred of words) and from specific (e.g. the \emph{Star Wars} universe or the O. J. Simpson murder case) to general (e.g. demography or elections). 
In the case of the 20 Newsgroups corpus (18K newsgroups emails), SMH discovered several topics that loosely correspond to the main thematic of the different newsgroups from where the documents were collected.
For example, the sample topics from 20 Newsgroups in Table~\ref{tbl:examples} are related to religion, computers, sports, cryptography, politics, space and medicine. 
On the other hand, most topics from the Reuters corpus (800K news articles) are related to major world events, important world news, economy, finance, popular sports and technology; the Reuters sample topics in Table~\ref{tbl:examples} are related to the stock market, elections, football, the O. J. Simpson murder case, Microsoft Windows, the Mexican economy, and the mad cow disease.
Finally, a wide variety of topics were discovered from both Spanish and English editions of Wikipedia, including demography, history, sports, series, and music.
It is also worth noting the similarity of some discovered topics that appear in both English and Spanish Wikipedia, e.g. the sample topics related to demography, the \emph{Star Wars} universe, and American football.

\section{Conclusions}
\label{sec:conclu}
We presented Sampled Min-Hashing (SMH), a simple approach for automatically discovering topics from collections of text documents which leverages Min-Hashing to efficiently mine and cluster beyond-pairwise word co-occurrences from inverted file bags.
This approach proved to be highly effective and scalable to massive datasets, offering an alternative to topic models which have dominated topic discovery for over a decade.
Moreover, SMH does not require a fixed number of topics to be determined in advance.
Instead, its performance depends on the inherent co-occurrence of words found in the given corpus and its own parameter settings.
We showed that SMH can discover meaningful and coherent topics from corpora of different sizes and diverse domains, on a par with those discovered by Online LDA.
Interestingly, the topics discovered by SMH have different levels of granularity that go from specific (e.g. a topic related to a particular capital stock) to general (e.g. a topic related to stock markets in general).

In SMH, Min-Hashing is repurposed as a method for mining highly co-occurring word sets by considering each hash table as a sample of word co-occurrence.
The coherence of the topics discovered by this approach was stable over a range of parameter settings.
In particular, our experiments demonstrated the stability of SMH over different values of the $JCC_B$ threshold $\eta$, the tuple size $r$ and the overlap coefficient $\epsilon$.
In our evaluation we found that many interesting co-occurring word sets had smaller $JCC_B$ values, and thus posit that smaller $\eta$ values may improve recall albeit at a higher computational cost.
Similarly, larger tuple sizes $r$ may improve precision at a very high computational cost.
Our empirical results suggest that a tuple size of $r = 2$ with a $JCC_B$ threshold of $\eta = 0.04$ provides a good trade-off between recall, precision and efficiency.
With those parameter values, maximum coherence is achieved when setting the overlap coefficient threshold to $\epsilon = 0.9$ without compromising speed or recall.
Finally, smaller vocabulary sizes tend to slightly improve topic coherence while considerably reducing recall.
In general, we found that SMH can produce coherent topics with large vocabularies at a high recall rate, showing its robustness to noisy and uncommon words.

We demonstrated the scalability of SMH by applying it to corpora with an increasing number of documents and vocabulary sizes. 
We found that the discovery time required by SMH grows linearly with both corpus and vocabulary size.
Remarkably, SMH performed topic discovery on the entire English edition of Wikipedia, which contains over 5 million documents and 1 million words, in at most 6.4 hours on relatively modest computing resources.
As opposed to Online LDA and other versions of LDA, SMH's discovery time is not directly affected by the number of discovered topics, but instead by the number of documents in the corpus, the vocabulary size, the $JCC_B$ threshold $\eta$, and the tuple size $r$.
On the Reuters corpus, which has more than 800,000 documents and 100,000 words, SMH was significantly faster than Online LDA and, when the number of topics was set to 400 and 600, its advantage was greater still.

A side effect of not having tight control over the number of topics is that SMH can discover thousands of topics on a wide variety of themes with different levels of granularity, including some overly specific that may be difficult to interpret if the user is not an expert in the given domain. 
However, we strongly believe that some of these specific topics are interesting and can provide deeper insights into the corpus.
Thus, more sophisticated visualization and browsing tools need to be developed in order to alleviate the burden of exploring and interpreting such a large number of diverse topics.   

Our current implementation of SMH requires the inverted file of the corpus as well as the CWS to be kept in RAM during the discovery process.
Because of this, when the corpus grows large, memory requirements may become high for a single computer.
In addition, our implementation is sequential and does not take advantage of multicore processors or distributed systems.
However, given that each hash table can be computed independently, SMH is highly parallelizable.
In the future, we plan to develop a parallel and distributed version of SMH which could scale up to even larger corpora and make the discovery process much faster.
An online version of SMH where discovery can be made incrementally with small batches of the corpus would also help reduce memory requirements and allow SMH to be applied at streams of documents.

The results presented in this paper provide further evidence of the relevance and generality of beyond-pairwise relationships for pattern discovery on large-scale discrete data.
 This opens the door for several possible directions for future work. 
 For example, it would be interesting to explore the use of SMH topics as a prior for LDA or other topic models. 
 Another interesting research direction would be to devise a general sampling scheme based on SMH for such models.  
Furthermore, the proposed approach could be useful in other problem domains where recurrent patterns are immersed in massive amounts of data, such as bioinformatics and astronomy.
 
\section*{Acknowledgements}
We acknowledge the resources and services provided for this project by the High Performance Computing Laboratory (LUCAR) at IIMAS-UNAM.
We also thank Adrian Durán Chavesti for his invaluable help while we used LUCAR, Derek Cheung for proofreading the manuscript and the editor and anonymous reviewers for their valuable comments that help improve this manuscript.

\bibliographystyle{model5-names}
\biboptions{authoryear}
\bibliography{references}
\end{document}